# A Stratified Simulation Scheme for Inference in Bayesian Belief Networks

**Remco R. Bouckaert**– PhD student
Utrecht University
Department of Computer Science
P.O.Box 80.089 3508 TB Utrecht, The Netherlands
remco@cs.ruu.nl

## Abstract

Simulation schemes for probabilistic inference in Bayesian belief networks offer many advantages over exact algorithms; for example, these schemes have a linear and thus predictable runtime while exact algorithms have exponential runtime. Experiments have shown that likelihood weighting is one of the most promising simulation schemes. In this paper, we present a new simulation scheme that generates samples more evenly spread in the sample space than the likelihood weighting scheme. We show both theoretically and experimentally that the stratified scheme outperforms likelihood weighting in average runtime and error in estimates of beliefs.

## 1 Introduction

Simulation schemes [Chavez and Cooper, 1990; Henrion, 1988; Pearl, 1992; Shachter and Peot, 1990] offer simple and general-purpose procedures for inexact probabilistic inference in Bayesian belief networks. The basic idea underlying these schemes is to generate a set of samples and to approximate beliefs of various variable values by the frequency of appearance in the sample.

Exact inference in Bayesian belief networks has been proven NP-hard, [Cooper, 1990]; so, exact algorithms [Lauritzen and Spiegelhalter, 1988; Pearl, 1988; Shachter, 1988] all have an exponential in the number of variables complexity. Though when demanding a certain accuracy in beliefs, runtimes of simulation schemes are also NP-hard [Dagum and Luby, 1993] and an exponential in the number of nodes amount of samples is necessary, the runtime is linear in the number of samples and variables.

The complexity of exact methods strongly depends on the topology of the network; especially when many loops occur in a network, the performance of exact methods decreases dramatically. However, for simulation schemes the topology of the network does not matter. In many applications exact inference may not be necessary since, due to inexactness of the probability assessments in the network, approximate beliefs suffice.

However, observation of values of variables and propagation of this evidence tends to decrease the performance of simulation schemes; many samples may be very non-specific for the observed situation and only a small portion of the samples may influence the estimates of beliefs. Therefore, it is important that a simulation scheme generates a lot of samples evenly distributed over the sample space. Generating such samples in an efficient way is the topic of the present paper.

In Section 2, we review some of the most popular simulation schemes in a general framework. In Section 3, we present a new scheme based on a popular statistical technique called stratification. The complexity and several optimizations of this scheme are described. In Section 4, we present experimental results comparing various simulation schemes. We end with conclusions in Section 5.

## 2 Simulation Schemes for Bayesian Belief Networks

Let $U = \{x_1, \ldots, x_n\}$, $n \geq 1$, be a set of variables; for simplicity we assume the variables are discrete. A *Bayesian belief network* $B$ over $U$ is a pair $(B_S, B_P)$ where the *network structure* $B_S$ is a directed acyclic graph with one node for each variable in $U$. $B_P$ is a set of conditional probability tables. For every variable $x_i \in U$, the set $B_P$ contains a conditional probability table $P(x_i|\pi_i)$ that enumerates the probabilities of all values of $x_i$ given values of the variables in its *parent-set* $\pi_i$ in the network structure $B_S$. The probability distribution represented by such a belief network $B$ is $\prod_{x_i \in U} P(x_i|\pi_i)$, [Pearl, 1988].

Let $E$ be the set of values of observed variables. Inference in a belief networks amounts to calculating the beliefs $Bel(x)$ in each variable $x$, that is the probability of the values of each variable given $E$, $P(x|E)$. Simulation schemes aim to approximate these beliefs by randomly generating samples. A *sample* is a value assignment to all variables in $U$, also called *instantiation*. The scheme keeps track of the relative frequency of variable values in the samples called the *score*.

In Figure 1, a general sampling algorithm is depicted which we will use as a general framework to describe various simulation algorithms. Depending on the method of sample generation, an initialization procedure is executed. Then, $m$ samples are generated and for each generated sample $S$, the quotients of the probability of the instantiation, $P(S)$, and the probability of generating the instantiation, $P(\text{selecting } S)$, is calculated. With this value, the score is updated. Eventually, the scores are normalized to obtain the beliefs.

First, we concentrate on methods for generating samples and initialization, and turn to scoring methods shortly. Henrion [Henrion, 1988] introduced a sampling algorithm for belief networks. The value assignments of the separate variables are chosen equiprobable; the probability of selecting an instantiation therefore is equal for all instantiations. No initialization is performed for this scheme. A slight optimization is to generate only values for the variables for which no evidence has been obtained. These variables get assigned their observed value in each sample. The value of $p$ is calculated by $\prod_{x_i \in U \setminus E} P(x_i|\pi_i)$. We call this scheme the *simple* scheme.

Another method of sample generation was proposed in [Henrion, 1988]. First, values for the root nodes of the network are generated with probabilities equal to the probabilities of the probability table first. Then, for the nodes of which all parents have been assigned a value values are generated with probabilities equal to the chance of these nodes given the values assigned to their parents. For this procedure it is handy to have a topological ordering on the variables which needs to be calculated during initialization. Again, evidence nodes are assigned their observed values in each sample. The value of $p$ is calculated by $\prod_{x_i \in E} P(x_i|\pi_i)$. We call this method *likelihood weighting*. This method is also known as logic sampling [Henrion, 1988] and evidence weighting [Fung and Chang, 1990; Shachter and Peot, 1990].

The last method considered here was proposed by Pearl [Pearl, 1992] which relies on Markov blankets. The *Markov blanket* $Bl(x_i)$ of a node $x_i$ consists of the parent-set of $x_i$, the children of $x_i$ and the parents of these children except for $x_i$ itself. In this method, a sample is not generated independent of the previous samples. When generating a new sample the previous sample is taken into account: the new value of a node $x$ is chosen with probability proportional to the product of probabilities in its Markov blanket $Bl(x)$, $\prod_{x_i \in Bl(x)} P(x_i|\pi_i)$. Note that the probability of selecting a sample $S$ is $P(S)$, so $p = 1$. As in the other methods, evidence nodes are assigned their observed value. We call method procedure *Pearl's scheme*.

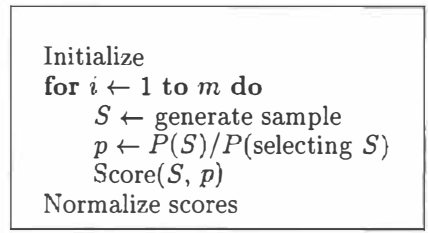

Figure 1: General Sampling Algorithm.

We consider two scoring methods; simple scoring and Markov blanket scoring. *Simple scoring* is done by adding the value $p$ yielded by the sample generating method for a sample $S$ to the score of each variable with the value it has in $S$. A more effective approach [Pearl, 1992] seems to add $p$ to every value of the variable weighted by the probability of its Markov blanket. The latter method will be called *Markov blanket scoring*. For the simple and likelihood weighting scheme extra work needs to be done when Markov blanket scoring is used namely the calculation of the product of the probabilities over the Markov blankets. However, for Pearl's scheme these probabilities are already available, so little extra work needs to be performed for this scheme.

## 3 A Stratified Simulation Scheme

Stratification is a popular statistical technique for obtaining samples that are more uniformly spread in the sample space. A description can be found in any basic book on sampling. The basic idea is to divide the sample space into so-called strata, and choose in each stratum a given number of samples. Such samples represent the distribution better than randomly chosen samples, because it is not possible that no samples are taken from a large area of the sample space. So, less samples are required for a similar error in estimates. There is a large freedom in selecting strata. In our approach, we will split the sample space into $m$ equally likely strata an choose one sample from each stratum. As in Pearl's scheme, we allow some dependence among samples. This dependence makes it possible to generate the samples faster than in the simple, the likelihood weighting and Pearl's scheme.

### 3.1 Stratification for Bayesian Belief Networks

Let the variables in $U$ be ordered $x_1, \ldots, x_n$. For ease of exposition, assume all variables to be binary taking values from $\{0, 1\}$. Then, instantiations of $U$ can be ordered according to $0 < 1$ taking order of variables in account. With each instantiation $S$ of $U$ we associate an interval $I(S)$ defined by,

$$I(S) = [lo(S), hi(S)),$$

where $lo(S) = P(U < S) = \sum_{S' < S} P(S')$, and $hi(S) = lo(S) + P(S)$. The unit interval is divided into subintervals and every instantiation of $U$ is assigned such a subinterval. Alternatively, every number $r$ in the unit interval corresponds to an instantiation $S$ of $U$ such that $r \in I(S)$. For example, let $U = \{a, b, c\}$ and let $P(U)$ be defined by the Bayesian belief network depicted in Figure 2. Suppose that the variables are ordered $a, b, c$, we have for the values $a = 0$, $b = 1$, and $c = 0$, that is instantiation $S = 010$, the interval $I(010)$ associated with $S$ which is $[0.15, 0.325)$. Since $lo(S)$ is $P(000) + P(001) = 0.15$ and $hi(S)$ is $P(S) = 0.5 \times 0.7 \times 0.5 = 0.175$ plus $lo(S)$ which equals $0.325$.

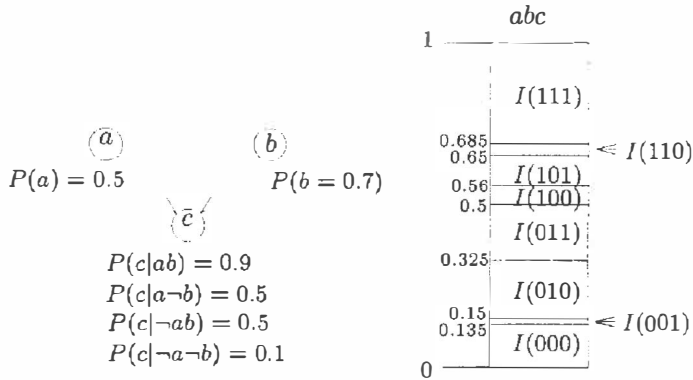

Figure 2: Belief network and corresponding intervals.

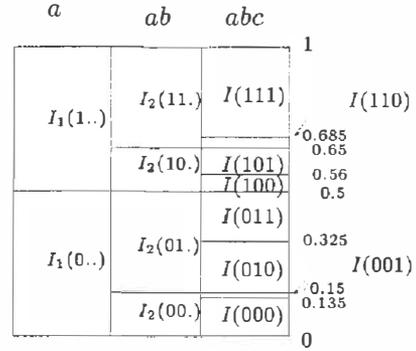

Figure 3: Intervals of prefixes.

The stratified simulation scheme is based on using these intervals to determine samples. In its simplest form, a number $r$ is randomly chosen from the unit interval and the instantiation corresponding to the interval that includes $r$ is a sample generated. In our example, suppose that the number $r = 0.2345$ is chosen. Then, $r$ is in the interval $[0.15, 0.325)$ corresponding to instantiation $S = 010$. So, the sample $a = 0$, $b = 1$ and, $c = 0$ is generated.

By imposing certain restrictions on the number chosen from the unit interval, a more efficient simulation scheme is yielded. Suppose $m$ random numbers are chosen in the unit interval and these numbers then are considered in ascending order. Now suppose that the numbers $r_1 = 0.2345$, $r_2 = 0.4567$, and $r_3 = 0.6789$ have been generated. The sample corresponding to the first number is $S_1 = 010$, to the second $S_2 = 011$, and to the third $S_3 = 110$. Observe that for the samples $S_1$ and $S_2$ only the least significant bit has changed. In general, when the random numbers are considered in ascending order, then only the $k$ least significant bits change and the $n - k$ most significant bits do not. This property can be exploited to get a more efficient simulation scheme. We only have to put computational effort in assigning values to these least significant variables, while in the other simulations schemes, all variables need to be updated. However, we need to do some extra work to determine which variables need to be updated. To do so, we generalize the definition of intervals to apply to prefixes of instantiations.

Let $pref_k(S)$, $0 \le k \le n$, be the prefix of $k$ bits of instantiation $S$. So, $pref_3(0111)$ is $011$ and $pref_1(0111)$ is $0$. Then, the intervals generalized to prefixes $I_k(S)$ associated with instantiation $S$ is defined by,

$$I_k(S) = [lo_k(S), hi_k(S)),$$

where $lo_k(S) = P(pref_k(S') < pref_k(S)) = \sum_{pref_k(S') < pref_k(S)} P(S')$ and $hi_k(S) = lo_k(S) + P(pref_k(S') = pref_k(S)) = \sum_{pref_k(S') \le pref_k(S)} P(S')$. Note that for $k = n$ we have the original definition for intervals, that is $I_k(S) = I(S)$, and for $k = 0$, we have the entire unit interval, $I_0(S) = [0, 1)$. Figure 3 shows the intervals for our example; $I_2(01.)$ starts at $0.15$ since $P(pref_2(S) < 01) = P(000) + P(001) = 0.15$ and ends at $0.5$ since $P(pref_2(S) = 01) = P(010) + P(011) = 0.35$. Also from this definition follows that $I_k(S) \subseteq I_{k-1}(S)$.

Therefore, when we are looking for an interval that contains $r_i$ and we the previous sample is instantiation $S_{i-1}$, first we check if $r_i$ is in $I_n(S_{i-1})$. If it is not, we check if it is in $I_{n-1}(S_{i-1})$ and so forth, until we find a $k$ such that $r_i$ is in $I_k(S_{i-1}) = [lo_k(S_{i-1}), hi_k(S_{i-1}))$. Now observe that for all $j$, $lo_j(S_{i-1})$ is smaller than $r_i$. So, only $hi_j(S_{i-1})$ need to be considered; looking for $k$ such that $hi_k(S_{i-1}) > r$ and $hi_{k+1} < r$ is sufficient. Since $hi_k(S_{i-1})$ is a descending function of $k$, this procedure can be performed with binary search, which costs at most $\log n$ operations (all logarithms in this paper are to base 2 unless stated otherwise). Note that this procedure easily generalizes to non-binary variables.

However, we will not generate numbers randomly in the unit interval and then consider them in ascending order. Instead, we divide the interval into $m$ equal

```
l₀ ← 0;  h₀ ← 1
for  i ← 1 to n do
    lᵢ ← 0
    if xᵢ ∈ E then
        valᵢ ← eᵢ
        hᵢ ← hᵢ₋₁
    else
        valᵢ ← 0
        hᵢ ← hᵢ₋₁ * P̃ᵢ(0)
```

Figure 4: Initialize Stratified Scheme.

```
f ← (random[0 : 1] + i − 1)/m
j ← Binsearch (f, h)
while j <= n do
    if xⱼ ∈ E then
        lⱼ ← lⱼ₋₁
        hⱼ ← hⱼ₋₁
    else
        k ← 0
        lⱼ ← lⱼ₋₁
        hⱼ ← lⱼ + (hⱼ₋₁ − lⱼ₋₁) * P̃ⱼ(k)
        while f > hⱼ do
            k ← k + 1
            lⱼ ← hⱼ
            hⱼ ← lⱼ + (hⱼ₋₁ − lⱼ₋₁) * P̃ⱼ(k)
        valⱼ ← k
    j ← j + 1
return(val)
```

Figure 5: The $i$th Sample Generation Method for the Stratified Scheme.

strata where $m$ is the number of required samples, and for each stratum we generate one random number $r_i$. This procedure guarantees that the samples are uniformly chosen from the sample space.

### 3.2 An Algorithms for the Stratified Scheme

Based on these observations we formulate a stratified scheme for generating samples that fits in the general algorithm shown in Figure 1 of the previous section. The strata are regarded in ascending order. In each stratum a number $r$ is randomly chosen. For that stratum, dynamically a new instantiation and new intervals are calculated. We need to define initialization and sample generation methods. In Figure 4 and 5, pseudo-code for these methods is shown. The values of the variables for a sample is stored in the array $val$. We keep track of the intervals in the arrays $l$ and $h$ for respectively the lower and upper bound of the intervals of the instantiation stored in $val$. For initialization, an instantiation $S_0$ is generated in which the value of each variable is set to 0 except when there is evidence for the variable. Obviously, the lower bounds of the intervals are 0 initially, that is $lo_j(S_0) = 0$, $0 \leq j \leq n$. The upper-bound $hi_j(S_0)$ is the upper-bound of the previous interval $h_{j-1}(S_0)$ times the probability $P̃_i(0)$ of choosing the value of variable $x_i$. There are several ways of defining $P̃$. When $P̃_i$ is chosen the reciproce of number of values $x_i$ can take, all states are equiprobable and this scheme will be referred to as the *stratified simple scheme*. However, one can also take for $P̃$ the probability of choosing that value of variable $x_i$ given its parent as instantiated in $val$. This scheme will be referred to as the *stratified likelihood scheme*. Note that evidence nodes do not contribute to the sample.

Figure 5 shows pseudo-code for the method for generating a sample. First, a random number $r$ in the $i$th section is generated. Using binary search, the first variable $x_j$ for which $h_j < r$ and $h_{j-1} > r$ is identified. For the variables $x_j$ up to $x_n$ a new value will be calculated while the values $val_1$ up to $val_{j-1}$ remain unchanged. The boundaries of the intervals of $x_j$ are calculated from the boundaries of $x_{j-1}$; if $x_j$ is an evidence node then the boundaries are the same as for $x_{j-1}$. If $x_j$ is not an evidence node then they are bounded by the boundaries of $x_{j-1}$. The value of variable $x_j$ is calculated by stepping through the range of $x_j$ until the boundary encloses $r$.

### 3.3 Performance of the Stratified Scheme

Now, we consider the amount of work that needs to be performed for generating $m$ samples in a belief network with $n$ nodes. When generating a new sample, our scheme saves the work of determining values for $k$ variables at the cost of at most $\log n$ comparisons. We investigate the computational complexity of our scheme in further detail. Suppose all variables are binary. Then, the most significant non-evidence variable gets assigned a value at most twice by our scheme, the second most significant non-evidence variables at most four times, etcetera; the $\lfloor \log m \rfloor$th up to the $n$th. Less significant non-evidence variables all get assigned a value at most $m$ times because they cannot get $2^{\lfloor \log m \rfloor + k}$ $(k > 0)$ times an assignment in $m$ samples. So, at most

$$\sum_{i=1}^{\lfloor \log m \rfloor} 2^i + (n - \lfloor \log m \rfloor - 1).m,$$

variable assignments are performed. At most $m$ times a binary search is performed. Using $\sum_{k=0}^{x} 2^k = 2^x - 1$ and including the binary searches, we find that sample generation involves at most,

$$2^{\lfloor \log m \rfloor} - 2 + (n - \lfloor \log m \rfloor - 1)m + m \log n,$$

operations. We find that our scheme has a computational complexity of order $O((n - \frac{\log m}{n})m)$. When the arity of the variables is at most $k$, this becomes $O((n^{-k} \log \frac{m}{n} + \log^2 n)m)$. Note that if the number of samples $m$ is larger than the number of variables, the stratified simple and likelihood weighting scheme are more efficient than the simple and likelihood weighting schemes which are of complexity $O(n.m)$.

We conclude our analysis by observing that the complexity bound is conservative; if probabilities of the lower ordered variables are close to one and the stratified likelihood weighting scheme is used, a much smaller number of samples is chosen for stratified schemes. It is assumed that the work for one comparison in a binary search is equally expensive as determining the value of a variable; however, for determining the value of a variable $x$, its probability table need to be looked up which may be relatively expensive if $x$ has many parents.

In general, estimates of beliefs become more accurate when the number of samples increases. Dagum and Horvitz [Dagum and Horvitz, 1993] showed that for the likelihood weighting scheme, to output a belief in a value of a variable $x$ that with probability higher than $1 - \delta$ has relative error less than $\epsilon$, at least $a \cdot \ln(4/\delta)/(\epsilon^2 Bel(x))$ samples are required where $a$ is the maximum value of the weighting distribution. Consider once more the example of Figure 2. For even numbers of samples, always an equal number of samples with $a = 0$ and with $a = 1$ will be generated. This results in a correct estimate of the probability of $a$ namely $P(a) = 1/2$. So, the algorithm also produces better samples, a point stressed in [Chavez and Cooper, 1990] to be very important. Especially for variables that are low in the ordering good samples are produced. I feel that the bound of Dagum and Horvitz may be taken as upper bound to the number of samples to be generated.

### 3.4   Further Optimizations

The previous section presented a new sample scheme that is shown to be faster than other popular sampling schemes. In this section, we give attention to details of the scheme in order to get a better performance.

It is desirable to generate many samples in a small amount of time. To do so, it is important to choose the data-structures to be used carefully. Since conditional probability tables are accessed very often, we focus on the data-structure to store these tables. These tables may be stored in an array; the basic idea is illustrated in Figure 6 for the probabilities from the tables of variable $c$ of the example of Figure 2. For example, in CABeN [Cousins $et$ $al.$, 1991], a collection of algorithms for belief networks, probability tables are implemented this way. Note however that to access the array an index needs to be calculated from the instantiation of the parents of this variable. The calculation of such an index requires computationally expensive multiplications. If the network contains binary variables only however, the multiplications can be replaced by shift operations. In experiments on a HP-9000 series 700 using a C-program using shift op-erations instead of multiplication for calculating the array index resulted in a 15% reduction of computer time.

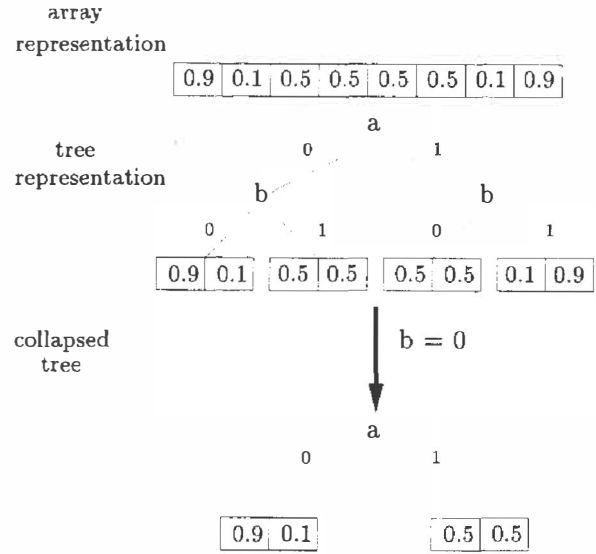

Figure 6: Storing conditional probability table of node $c$.

Instead of arrays, search trees offer an alternative data structure for storing probability tables. A search tree is a tree in which on a node a choice is made which branch to take and the leafs contain information. In Figure 6 such a search tree for the probability table of the variable $c$ from Example 6 is depicted. To find the required probability, only a pointer needs to be passed through the tree and no multiplication is performed. In experiments on a HP-9000 series 700 using a C-program using search trees instead of arrays to store probability tables resulted in a 30% reduction of computer time.

The search tree also offers other advantages. When evidence is observed, outgoing arcs of the observed nodes can be removed [Gaag, 1993] and the probability tables can be collapsed; the idea is that if variable $x$ is observed to be 1 then all children of $x$ will not use probabilities conditioned on instantiations in which $x$ is not 1. Therefore, those probabilities can be removed from the probability table. To implement this, a search tree that stores the probability table can be pruned; only those leaves in the search tree for which the observed value is present need to be stored. This is an almost trivial action for trees while it would require considerable computing for arrays. For example when $b$ is observed to be 0, the search tree for the representation of the probability table of $c$ can be replaced by the lower tree depicted in Figure 6.

Not only the choice of data structures is important for optimal performance. The stratified likelihood weighting scheme needs a topological ordering on the variables. Such an ordering is not unique. To fully exploit the reduction in time achieved by the stratified scheme, variables with high probabilities should occur foremost in the ordering; in that case they won't need a change of value too often. Therefore, when determining a topological order of the variables, their probability tables should be taken in consideration. In our experiments, we used the average probability to the power four $\sum_{x_i, \pi_i}^{*} P(x_i|\pi_i)^4 / \sum_{x_i, \pi_i} 1$ as an extra criterion to sort the variables since it assigns extra weight to probabilities close to one; small probabilities vanish while large probabilities contribute a lot to this sum. However, we think it is worth to investigate other criteria. Since when evidence is observed, outgoing arcs of the observed nodes can be removed, less constraints are left for choosing a topological ordering; children of observed nodes may be shifted lower in the ordering if their probabilities are high enough.

So far, we assumed that a random number in each section is chosen. However, also the median of the interval can be taken. At least for the lower ordered nodes, no change in estimates are expected. In fact, these estimates will become better because less errors due to random fluctuations are introduced. For variables high in the ordering however, it has the same effect as choosing a random number.

Care must taken when networks with many variables are used; the values of $lo_k(S)$ and $hi_k(S)$ may be erroneously calculated as equal due to numerical round off errors. Therefore, the representation size used for $lo_k(S)$ and $hi_k(S)$ need to be taken large enough. If a random number is chosen from a section, also this random number must have enough precision to avoid biases.

## 4 Experimental Results

We have performed some experiments to compare the stratified simulation scheme with the simple scheme, likelihood weighting and Pearl's scheme. We generated randomly ten belief networks with fifty binary variables and a poly-tree structure. The networks were generated by ordering the variables, randomly pick two nodes $a$ and $b$ and adding the arc $a \to b$ if $a$ is lower ordered than $b$. Otherwise the arc $b \to a$ is added. This step is repeated but now one variable is randomly chosen from the variables that are connected to at least one arc and one variable is chosen from the variables that are connected to no arc. This last step is repeated till all arcs are placed.

With these ten networks we applied the four algorithms generating 100 up to 1000 samples, increasing by 100 in each test, and further, from 1000 with steps of 1000 up to 10000 samples. So, with every network 19 different sets of samples were generated. The probability tables were stored in search trees as described in Section 3.4. The performance of the algorithms was measured in time in milliseconds used to execute the algorithm according to the UNIX time-function. Furthermore, we judged the quality of the approximated beliefs by the divergence, that is the average logarithm of the estimated belief and real belief, $1/|U| \sum_{u \in U} P(u) \sum_{u \in \{0,1\}} \log(P(u)/\hat{P}(u))$. For simplicity, no evidence was used in the belief network.

Figure 7 shows the results for the simple scheme (simple), likelihood weighting (likelihood), Pearl's (pearl) and, the stratified schemes for both the simple (strat.s) and likelihood weighting (strat.l) variant. For all schemes simple scoring was used. The ordering of the variables was the same as the order used to generate the networks; the probability tables were not considered for the ordering. The closer the data-points are to the left lower corner, the better the performance of the scheme. The simple algorithm performed poorly and stratification does not really help. The reason for this behavior is that the samples chosen are mostly nonspecific for the distribution. Therefore, many samples are required to get a good performance and stratification does not influence this behavior very much. This is also expressed by the low slope of the data-points for the simple schemes. Likelihood weighting performed considerably better than Pearl's and the simple schemes, as was also reported in [Cousins et al., 1991; Shachter and Peot, 1990]. With the stratified likelihood weighting scheme even better performance is obtained, which was expected after the analysis in Section 3.

Figure 8 shows the results for the same algorithms as depicted in Figure 7 , this time using Markov blanket scoring. All data-points have shifted in the direction of the corner right under except for the points of Pearl's scheme. This could be expected since Markov blanket scoring results in much extra work for all but Pearl's scheme as pointed out in Section 2. So, the estimates become better at the cost of additional computational effort. Markov blanket scoring seems to help for Pearl's scheme.

Figure 9 shows the effects of incorporating various optimizations to the stratified likelihood weighting algorithm (strat.l): sorting the variables and using the extra criterion in the previous section (strat.l+s) and with random generation of numbers in a section versus taking median of the section (strat.l-r+s). The figure suggests that sorting helps but it helps only marginally. This could be expected since sorting with the extra criterion influences the order only marginally. The effect of using the median of a section instead of a random value does not seem to influence results though it makes the program simpler. Also this could be expected because it is only a minor adjustment of the algorithm. For a better comparison, the likelihood weighting algorithm (strat.l) is also depicted. For equal error levels, up to 30% less time is used by the best stratified scheme.

Figure 10 shows results for the best stratified algorithm, that is, with sorting and with taking the median of the section instead of a random value, with (strat.l-r+s+m) and without (strat.l-r+s) Markov blanket scoring. The figure suggests that Markov blanket scoring improves per test-set of samples the estimated probabilities yet takes extra time because of the additional computational effort that is required. These effects cancel each other out, so Markov blanket scoring does not seem to help but it also does no harm.

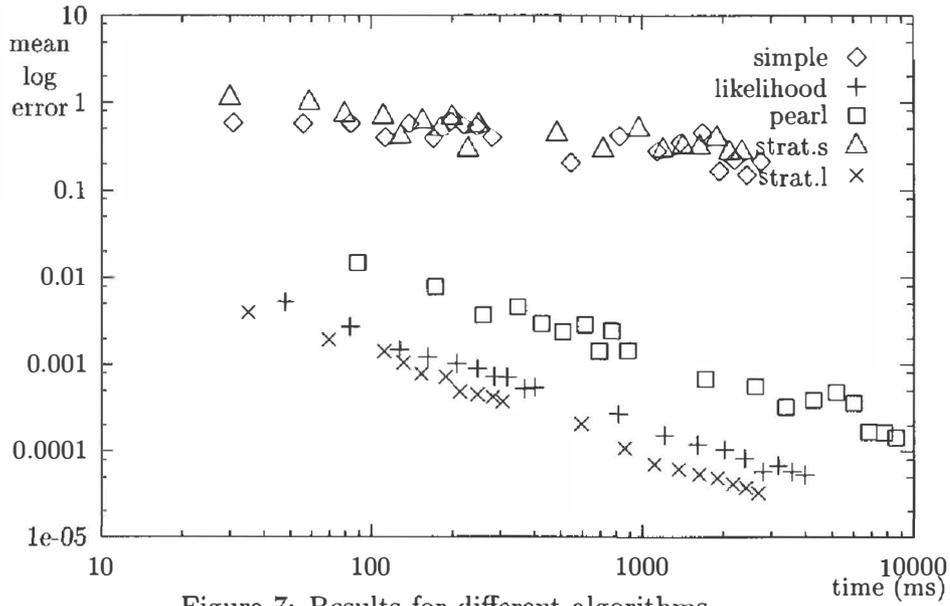

Figure 7: Results for different algorithms.

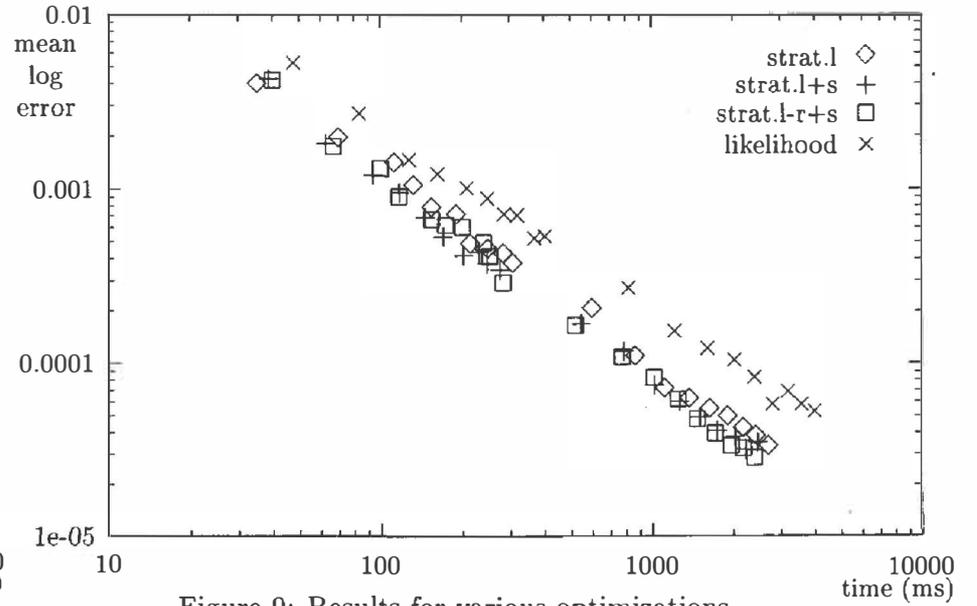

Figure 9: Results for various optimizations.

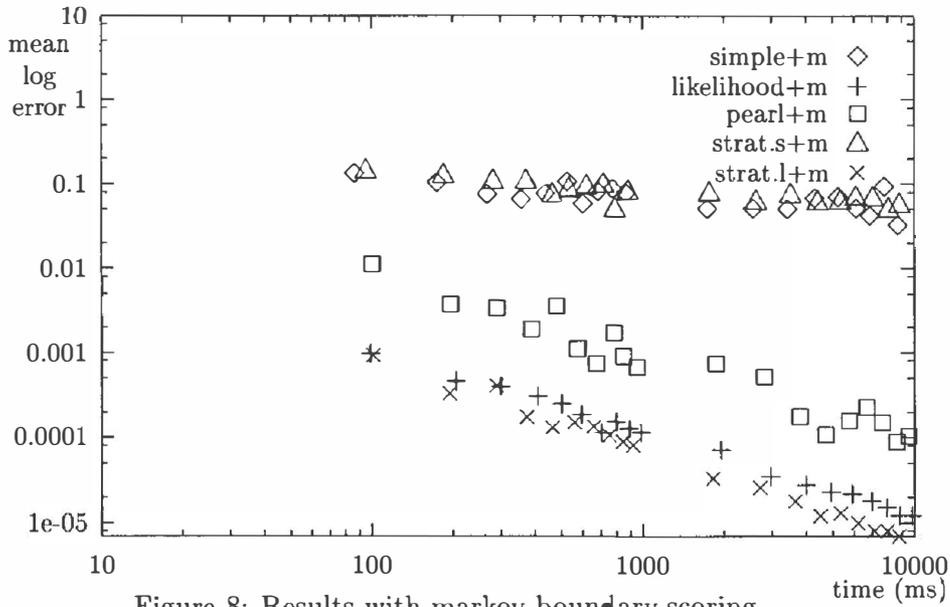

Figure 8: Results with markov boundary scoring.

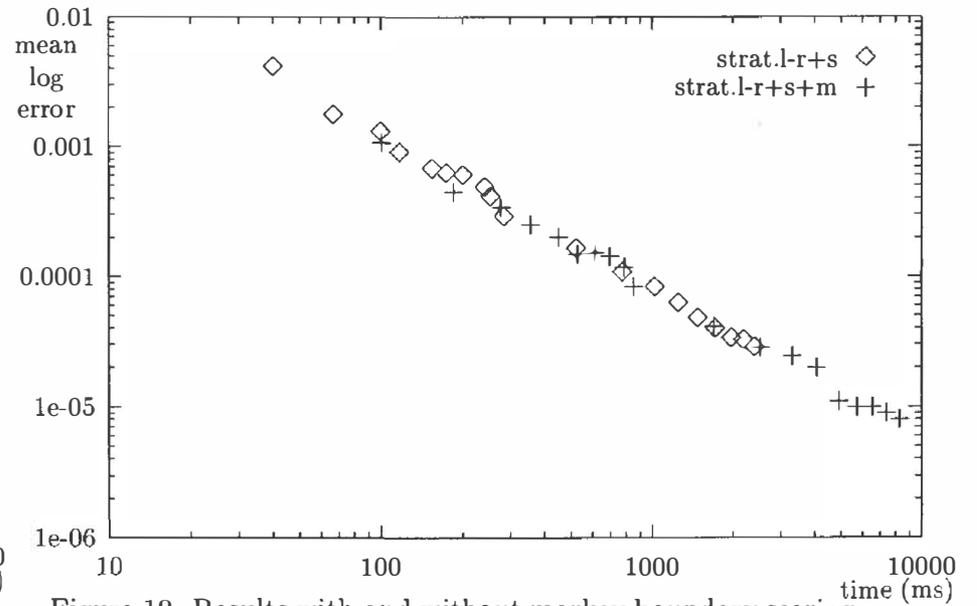

Figure 10: Results with and without markov boundary scoring.

## 5 Conclusions

In this paper, we presented a stratified simulation scheme for probabilistic inference in Bayesian belief networks. The scheme generates samples evenly spread in the sample space and can be implemented efficiently. The scheme is indeed more efficient than the likelihood weighting scheme. Due to the evenly spread samples, the scheme also result in better approximations of probabilities. We showed both theoretically and experimentally that approximation of beliefs is not only faster but also better than with existing schemes.

Though for special network structures exact algorithms may outperform simulation schemes, our algorithm offers a robust general purpose method for probabilistic inference without restrictions on the topology of networks.

The effects of various optimizations specific for the scheme were investigated. A variant where no random numbers are used performs equal to variants where random numbers are used. For the best variant of the stratified scheme, the extra computational effort necessary for Markov scoring cancels out the gain of better approximations of beliefs. The experiments have shown that some extra performance can be gained by choosing a clever ordering on the variables. Further research is necessary to investigate various sorting criteria on the performance of the algorithm.

### Acknowledgements

I thank Linda van der Gaag for her many useful remarks that improved the presentation of the paper and for the work of the anonymous referees.

# Proposal:
# Interactive Media for Research in Uncertainty

**Wray L. Buntine, RIACS**
NASA Ames Research Center, Mail Stop 269–2
Moffet Field, CA 94035–1000, USA
wray@kronos.arc.nasa.gov

## Introduction

Emerging forms of electronic media distributed via the Internet and the World Wide Web (WWW) have the potential to transform the way research and development is conducted. This proposal makes some suggestions for the development of an Internet resource for the uncertainty community in particular, and more generally for computational probability and decision theory.

Several computer science research communities maintain bibliographies, verified BIBTEX entries updated and distributed on a regularly basis (e.g., Computational Learning Theory, Computational Geometry). Other communities are building up Postscript libraries of theses, papers, and manuscripts, and WWW sites for conference abstracts, programs, etc. For example the Neuroprose Archive[1] in the connectionist community. This is successful because it is coupled with the Connectionist News Group moderated out of CMU, acting as a bulletin board and discussion group for new entries. Topics for workshops and emerging research areas are routinely born on this newsgroup, and technical reports contributed to the archive sometimes obtain a broad multidisciplinary feedback from from motivated readers, often higher quality than the subsequent journal reviewers.

Many believe these forms of interaction significantly improve the quality of publication, and the education and application of the research and development community. A considerably richer working environment can be developed with point-and-click WWW interfaces, which combines features such as:

- Browsing of distributed Postscript libraries.
- Menu/Forms driven remote operation of demonstration programs or bibliography servers.
- User authentication for automated reviewing, and controlled access to manuscripts.
- LaTeX, Word and Framemake to HTML (the mark-up language used by the WWW browsers)

translators that can allow rapid generation of basic material.

## The Proposal

A proposal for a basic Handbook for use by the community is outlined at the WWW site http://fi-www.arc.nasa.gov/ in the directory fia/users/buntine/Handbook/Overview.html (concatenate these two to get the URL). It is recommended that these be viewed as a suggestion rather than as guidelines, since any community development here would have to proceed in a growth path set by the community themselves.

A basic handbook format that I recommend the community adopt goes as follows:

- Establish a community bibliography maintained in BIBTEX and distributed in many formats, available for interactive browsing, etc. This would be maintained at a central location.
- Link the bibliography to distributed (e.g., author maintained) information about abstracts, key words, author details, errata, related papers (papers cited by, papers citing, etc.), so that the web of articles can be browsed.
- Link these in turn to a distributed Postscript library of the documents themselves housed at the authors FTP sites.
- Maintain news and notes recording relevant information such as pending conferences, call for papers, new books, tutorials, etc.

This basic system could subsequently be extended with tutorial and encyclopaedia entries linking into the bibliography.

## Acknowledgments

These ideas have been developed in cooperation with Bruce D'Ambrosio, Max Henrion Barney Pell, and Michael Frank.

---

[1]At archive.cis.ohio-state.edu, maintained by Jordan Pollack.